\documentclass[journal]{IEEEtran}
\usepackage{bm}
\usepackage{amsmath}
\usepackage{amssymb}
\usepackage{amsthm}
\usepackage{mathrsfs}
\usepackage{enumerate}
\usepackage{multirow}
\usepackage{color}
\usepackage{threeparttable}
\usepackage{subfigure}
\usepackage[square, comma, sort&compress, numbers]{natbib}
\usepackage[pagebackref=false,breaklinks=true,letterpaper=true,colorlinks,bookmarks=false]{hyperref}
\usepackage{times}
\usepackage{breakurl}
\usepackage{array}
\usepackage{verbatim}
\usepackage{algorithm}
\usepackage{bbding}
\usepackage{algpseudocode}
\usepackage{lettrine}
\definecolor{orange}{rgb}{1,0.5,0} %v2
\definecolor{purple}{rgb}{0.5,0,0.5} %v3

\makeatletter
\def\hlinewd#1{%
  \noalign{\ifnum0=`}\fi\hrule \@height #1 \futurelet
   \reserved@a\@xhline}
\makeatother

\ifCLASSINFOpdf
  \usepackage[pdftex]{graphicx}
  \usepackage{graphics}
  \usepackage{graphicx}
  \usepackage{epsfig}
  \usepackage{epstopdf}
  \graphicspath{{./fig/}}
  \DeclareGraphicsExtensions{.pdf,.jpeg,.png,.eps}
\else
  \usepackage[dvips]{graphicx}
  \graphicspath{{./fig/}}
  \DeclareGraphicsExtensions{.eps}
\fi

\begin{document}

\title{GiT: Graph Interactive Transformer for Vehicle Re-identification}
\author{Fei Shen, Yi Xie, Jianqing Zhu, Xiaobin Zhu, and Huanqiang Zeng
\thanks{This work was supported in part by the National Key R\&D Program of China under the Grant of 2021YFE0205400, in part by the National Natural Science Foundation of China under the Grants 61976098, 61871434, 61802136 and 61876178, in part by the Natural Science Foundation for Outstanding Young Scholars of Fujian Province under the Grant 2022J06023, and in part by Collaborative Innovation Platform Project of Fuzhou-Xiamen-Quanzhou National Independent Innovation Demonstration Zone under the grant 2021FX03 \emph{(Corresponding author: Jianqing Zhu and Huanqiang Zeng)}.}

\thanks{Fei Shen is with College of Engineering, Huaqiao University, Quanzhou, 362021, China and School of Computer Science and Engineering, Nanjing University of Science and Technology, Nanjing, 210094, China (e-mail: feishen@njust.edu.cn).}
\thanks{Yi Xie, Jianqing Zhu and Huanqiang Zeng are with College of Engineering, Huaqiao University, Quanzhou, 362021, China (e-mail: yixie@stu.hqu.edu.cn, e-mail: \{jqzhu and zeng0043\}@hqu.edu.cn).}
\thanks{Xiaobin Zhu is with School of Computer and Communication Engineering, University of Science and Technology Beijing, Xueyuan Road 30, Haidian District, Beijing, 100083 China (e-mail: zhuxiaobin@ustb.edu.cn).}
}

% The paper headers
\markboth{}%
%\markboth{Journal of \LaTeX\ Class Files,~Vol.~6, No.~1, January~2007}%
{Shell \MakeLowercase{\textit{et al.}}: Bare Demo of IEEEtran.cls for Journals}

% make the title area
\maketitle

\begin{abstract}

\color{black}
Transformers are more and more popular in computer vision, which treat an image as a sequence of patches and learn robust global features from the sequence. However, pure transformers are not entirely suitable for vehicle re-identification because vehicle re-identification requires both robust global features and discriminative local features. For that, a graph interactive transformer (GiT) is proposed in this paper. In the macro view, a list of GiT blocks are stacked to build a vehicle re-identification model, in where graphs are to extract discriminative local features within patches and transformers are to extract robust global features among patches. In the micro view, graphs and transformers are in an interactive status, bringing effective cooperation between local and global features. Specifically, one current graph is embedded after the former level's graph and transformer, while the current transform is embedded after the current graph and the former level's transformer. In addition to the interaction between graphs and transforms, the graph is a newly-designed local correction graph, which learns discriminative local features within a patch by exploring nodes' relationships.
Extensive experiments on three large-scale vehicle re-identification datasets demonstrate that our GiT method is superior to state-of-the-art vehicle re-identification approaches.

\end{abstract}


\IEEEpeerreviewmaketitle

%-------------------------------------------------
\section{Introduction}
\begin{figure}[tp]
    \centering
    \includegraphics[width=1\linewidth]{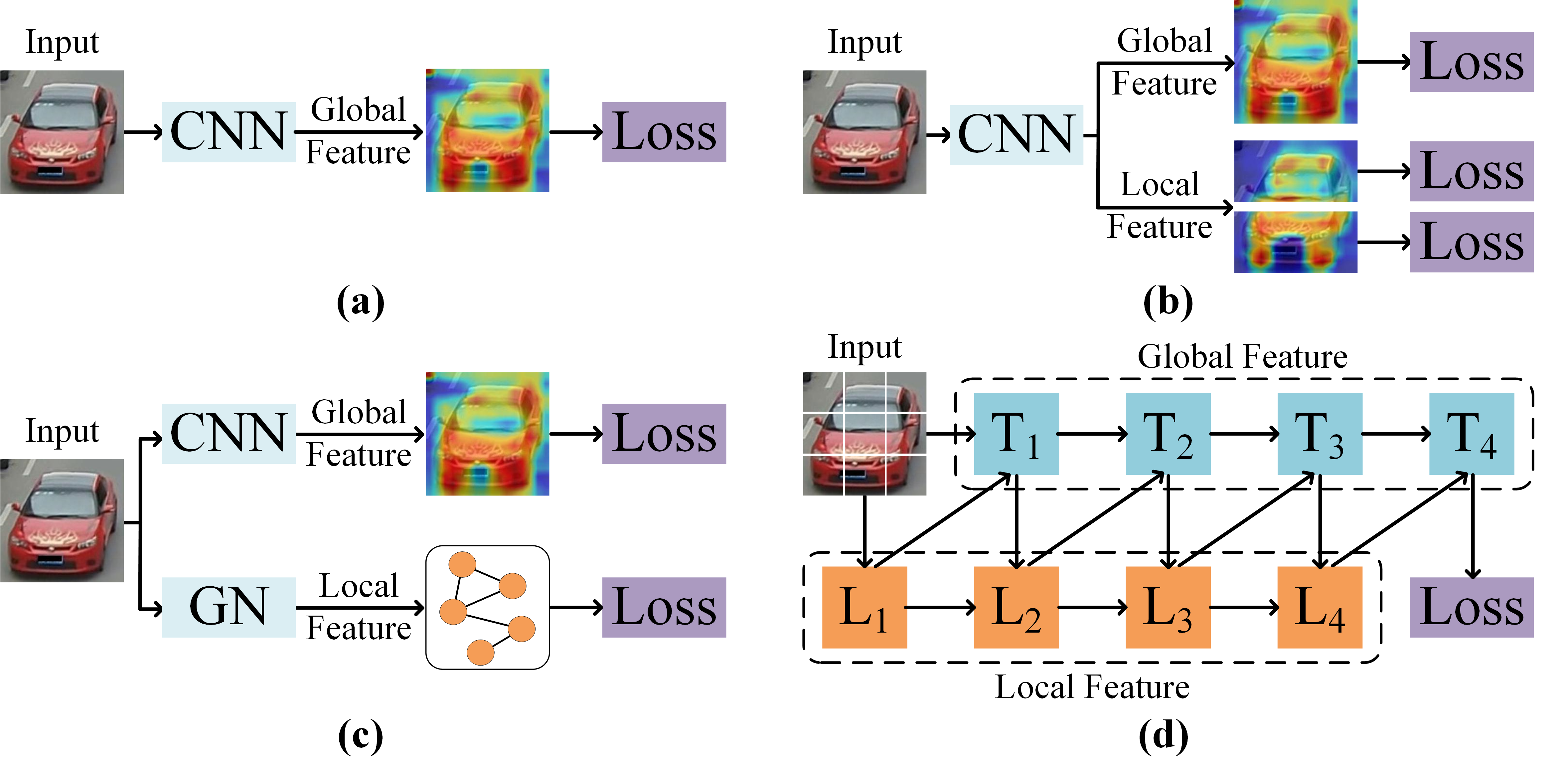}
    \vspace{-.3cm}
\caption{\small {Vehicle re-identification models use different architectures.
{\color{black}(a) Pure convolutional neural networks (CNNs) learn global features.
(b) Pure CNNs cooperate part divisions to learn global features and local features.
(c) CNNs combine graph networks (GNs) to learn global features and local features via a multi-branch structure.
(d) Our graph interactive transformer (GiT) method makes transformer (T) layers and local (L) correlation graph modules in an interactive status to learn global features and local features.}}}
%\vspace{-.2cm}
\label{fig:example}
\end{figure}

{\color{black}
%\vspace{-1.0cm}

\lettrine[lines=2]{V}{ehicle} {\color{black} re-identification \cite{guo2015nighttime, hsia2022new, guo2019two, lou2019embedding, liu2019group, zhou2018vehicle, shen2020net, veri2016, veri2017}
aims to search a target vehicle from an extensive gallery of vehicle images captured from different cameras, which has attracted much attention in the multimedia and computer vision community. }

With the widespread use of intelligent video surveillance systems, the demand for vehicle re-identification is growing exponentially.
{\color{black}
The two extrinsic factors hindering vehicle re-identification are poor illumination and various viewpoints. There are many vehicle re-identification studies \cite{li2020multi, aaver, pven, vanet, qddlf} focus on dealing with bad illuminations and pay attention to handling with various viewpoints. However, in addition to these two extrinsic factors, an intrinsic unfavorable factor is that vehicles of the same model and color are highly similar and difficult to be distinguished. The key to solving this inherent unfavorable factor is to treat subtle differences in wheels, lights, and front windows.
}
Therefore, effectively combining the global features and local features is crucial to improve the vehicle re-identification performance.

The development of vehicle re-identification technologies goes through three stages.
As shown in Fig. \ref{fig:example} (a), early methods \cite{ram, sdccnn, vst, dense, oife} apply pure convolutional neural networks (CNNs) to learn the vehicle images' global features. The architectures of these pure CNNs consist of a famous backbone network (such as VGGNet \cite{vgg}, GoogLeNet \cite{googlenet} and ResNet \cite{resnetv2}) and a global pooling layer.
Although early methods can learn global appearance features, they could not deal with local features, limiting the vehicle re-identification performance.

Vehicle re-identification methods enter into the second stage, which focuses on addressing global features' limitations. As shown in Fig. \ref{fig:example} (b), based on CNNs, there is a straightforward way of combining global features and local features learned from vehicle image partitions.
The partition division way can be further summarized into two kinds, i.e., uniform spatial division methods \cite{prn, qddlf, jquad, ram, san, sff, rpm, tamr} and part detection methods \cite{partreg, pgan, pmsm, oife, aaver, attdriven, pamtri}. The uniform spatial division methods do not require part annotations but are prone to suffer from partition misalignments. In contrast, the part detection methods can relieve dis-alignments but encounter a high cost of extra manual part annotations and massive training computations. No matter how to divide partitions, the subsequent feature learning is individually implemented on each part region, ignoring relationships among part regions.

Third, to consider relationships among part regions, vehicle re-identification methods enter the third stage, combining graph network (GN) with CNNs.
 As shown in Fig. \ref{fig:example} (c), in this stage, vehicle re-identification models \cite{liu_gcn, hssgcn, zhu2020graph, hpgn} usually have a CNN branch for learning global features and a GN branch for exploring the relationship among local features extract from part regions.
 {\color {black}However, down-sampling pooling and large-stride convolution operations reduce the resolution of feature maps, greatly restricting the ability of CNNs to recognize vehicles with similar appearances \cite{vit, vit-bot}.
 Besides, CNN and GN lack interactions as they are supervised with two independent loss functions, limiting vehicle re-identification performance.}

 {\color {black}
Recently, transformers \cite{vit, vit-bot, detr, transunet} have been attractive in computer vision. Transformers have two main advantages, namely, (1) global processing and (2) spatial information retaining. Regarding global processing, transformers use multi-head self-attention to capture global context information to establish long-distance dependence among patches neglected in CNNs. Regarding spatial information retaining, transformers discard down-sampling operations (i.e., 2-pixel stride convolution or pooling) that are usually required in CNNs. However, transformers also have disadvantages, as follows. (1) Transformers learning from patches violates the basic mechanism of human vision. (2) Transformers underrate the feature learning within patches, which simply uses fully connected or light convolutional layers to learn each patch's features. Overall, transformers achieve impressive performance boost over CNNs in many vision tasks \cite{segformer, track_transformer, hotr}. Especially, transformers obtain state-of-the-art performance on person re-identification \cite{vit-bot, hat, lai2021transformer}.

}

{\color{black}
{
	In this paper, we propose a graph interactive transformer (GiT) method for vehicle re-identification, as shown in Fig. \ref{fig:example} (d). The motivation is to let graphs learn local features within patches and transformers capture global relationships across patches and further couple graphs and transformers to cooperate local and global features effectively for improving vehicle re-identification.

}
}

{\color{black}
The main contributions of this paper can be summarized as follows:
\vspace{-.1cm}
\begin{itemize}
\item  We propose a graph interactive transformer (GiT) method to couple graphs and transformers , bringing an effective cooperation between local and global features. To the best of our knowledge, this paper is the first work that explores the interaction between local features and global features via graphs and transformers.

\item  We design a {\color{black}local correlation graph (LCG)} module for well learning discriminative local features from a patch, which subtly explores relationships of nodes within the patch.
The LCG module does not require any extra fine-grained part annotations or part division operations.

\item Experiment results on three large-scale vehicle datasets (i.e., VeRi776 \cite{veri776}, VehicleID \cite{drdl}, and VeRi-Wild \cite{wild}) show that the proposed method is superior to a lot of state-of-the-art vehicle re-identification approaches.
\end{itemize}
}

The rest of this paper is organized as follows. Section \ref{sec:rw} introduces the related work. Section \ref{sec:method} elaborates our method. Section \ref{sec:exp} presents the experimental results to show our method’s superiority. Section \ref{sec:con} concludes this paper.

\section{Related Work}\label{sec:rw}
In this section, we briefly review the most related works of vehicle re-identification and transformer in vision.
\subsection{Vehicle Re-identification}
{\color{black}
	%fully connection layer \cite{veri2016, drdl, mgr, imtri, djdl, abln, mattribute}
	%}
Convolutional neural networks (CNN) are commonly-used in vehicle re-identification. Hence, we review CNN models of vehicle re-identification from two aspects: (1) pure CNN-based methods and (2) combining CNN with graph network.

\textbf{Pure CNN-based methods.}
In recent years, a variety of pure CNN-based methods have been proposed for vehicle re-identification.
These methods usually include a robust CNN model for extract global features or local features and a feature aggregation structure for aggregate global features or local features.
According to the difference in feature aggregation structure, we can subdivide it into global feature learning methods and local feature learning methods.
The global feature learning methods \cite{veri2017, mgr, aaver, imtri, drdl,djdl, mattribute, vanet,vst,c2f,triemd,vric,msv,ealn,dban,vehiclenet} usually have a spatial global pooling layer to compress the entire vehicle features.
However, due to the characteristic of spatial global pooling layers, discriminative local features will inevitably be underestimated, which is detrimental to vehicle re-identification.

According to the local feature learning methods, we can further summarize it as uniform spatial division \cite{prn,qddlf,jquad,ram,san,sff,rpm,tamr} and part detection methods \cite{partreg,pgan,pmsm,oife,aaver,attdriven,pamtri}.
The uniform spatial division methods uniformly divide feature maps into several parts and then individually pools each region, as done in \cite{prn,qddlf,jquad}.
These methods are usually divided into several parts along the horizontal or vertical direction and then individually pool each region.
There is also a method of using a visual attention model to refine local features in \cite{sff, rpm}.
{\color{black}
 Although both vision transformer (ViT) and uniform spatial division methods learn attention on patches, the transformer uses multi-head attention to learn relationships among patches. The latter only uses self-attention to enhance local feature learning within patches. As a result, ViT can alleviate the problem of inaccurate division better than uniform spatial division methods.}
%However, due to vehicle appearance differences and camera viewing angle changes, these methods will face the problem of not accurately dividing the parts, limiting the performance of vehicle re-identification.

The part detection methods  \cite{partreg,pgan,pmsm,oife, hou2020detecting, aaver, attdriven,pamtri} can solve the problem of  inaccurate division.
These methods usually employ the typical detectors to detect vehicle parts or locate discriminative regions.
For example, the Part Regularization \cite{partreg} method uses you only look once (YOLO) \cite{yolo} as a detector to detect parts and feature extraction from part regions.
Two-level Attention network supervised by a Multi-grain Ranking loss (TAMR) \cite{tamr} detected the salient vehicle parts and proposed a multi-grain ranking loss to enhance the relationship of part features. Adaptive attention vehicle re-identification
(AAVER) \cite{aaver} employed key-point detection module to localizing the local features and use adaptive key-point selection module to learning the relationship of parts.
{\color{black} However, the main pain-point of part detection methods is that part detectors are deep networks in themselves, requiring a high cost of the extra manual part annotations, massive training, and inference computations.}

 \begin{figure*}[htp]
 	\centering
 	\includegraphics[width=.99\linewidth]{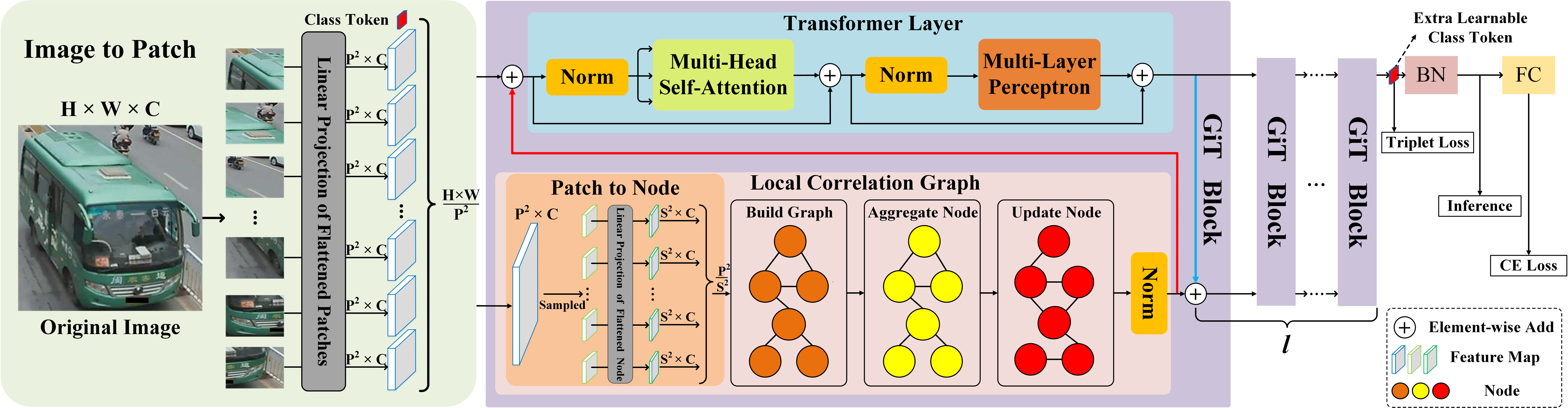}
 	%\vspace{-.25cm}
 	\caption{The framework of the proposed graph interactive transformer (GiT) method.
% The T and L denote a transformer layer and a local correlation graph (LCG) module.
 The BN and FC represent a batch normalization and a fully connected layer, respectively.}
 	\label{fig:framework}
 	 	%\vspace{-.2cm}
 \end{figure*}

}

\textbf{Combining CNN with Graph Network.}
Recently, various research efforts \cite{liu_gcn,zhu2020graph,hssgcn,hpgn,labnet,context,ddag,dgm,mace} combine CNN with graph network (GN)  for re-identification since GN can well extract local regional features. For example,  the parsing-guided cross-part reasoning network (PCRNet) \cite{liu_gcn} extracts regional features for each part from CNN and propagated local information among parts based on graph convolutional networks. The global structural graph \cite{global_structure} guides on features generated by a CNN to form fine-grained representations for vehicles. The structured graph attention network (SGAT) \cite{zhu2020graph} creates an inherently structured graph and an extrinsic structured graph to learn the vehicle's structure relationship feature. The hierarchical spatial structural graph convolutional network (HSS-GCN) \cite{hssgcn} enhances the spatial geometrical structure relationship among local regions or between the global region and local regions for vehicle re-identification.
{\color{black}
In addition to these vehicle re-identification works, there are several GN based person re-identification. The dynamic dual-attentive aggregation (DDAG) \cite{ddag} method reinforces representations with contextual relations crossing two modalities for visible-infrared person re-identification. For unsupervised video-based person re-identification, the dynamic graph matching (DGM) \cite{dgm} approach obtains a good similarity measure from intermediate estimated labels via an iteratively refining graph structure. The modality-aware collaborative ensemble (MACE) method \cite{mace} respectively handles modality-discrepancy in both feature and classifier levels by a middle-level sharable two-stream network for cross-modal person re-identification.
}

\subsection{Transformer in Vision}
The transformer is a type of self-attention-based neural network originally applied for natural language modeling (NLP) tasks.
Recently, attention-based transformers have gone viral in vision.
The vision transformer (ViT) \cite{vit} stacks multiple transformer layers.
Every transformer layer is a kind of residual structure. It is sequence packaged by layer normalization (LN), multi-head self-attention layer (MHSA) and a multilayer perceptron (MLP) block.
Based on the ViT models, many excellent networks are developed and verified for their effectiveness and scalability in downstream tasks.
For example, ViT-BoT \cite{vit-bot} combines the ViT framework with BNNeck \cite{luo2019strong} to construct a strong baseline for object re-identification.
Deformable transformer (DETR) \cite{detr} proposes multi-scale deformable attention modules to sample a set of critical points of patch for small objects detection.
{\color{black} The transformer version U-Net (TransUnNet)} \cite{transunet} takes the advantages of the CNN and transformer model to learn global context features for medical image segmentation.

Considering  the strong ability of the graph network to aggregate local features and the superiority of the ViT model to model global features,
we design a novel architecture to graph network and transformer layers are in a coupled status and demonstrate its effectiveness in this paper.
{\color{black}
Besides, there are also a similar topology method in multi-branch CNN, such as HRNet \cite{hrnet} and Multi-scale DenseNet \cite{ms_densenet}. All three methods both use a parallel structure in topology. However, our GiT is different from HRNet \cite{hrnet} and Multi-scale DenseNet \cite{ms_densenet}. Topologically, HRNet can two-way coupled multi-scale features, but the couple position is sparse. Multi-scale DenseNet \cite{ms_densenet} can densely couple multi-scale features, but the couple is single-way, i.e., from high-resolution features to low-resolution features. The GiT can conveniently achieve densely and two-way coupling of multi-scale features.
}

\section{Proposed Method}\label{sec:method}
\subsection{Overview}
As shown in Fig. \ref{fig:framework},  the proposed method's overall framework stacks a list of graph interactive transformer (GiT) blocks.
Each GiT block contains a newly designed local correlation graph (LCG) and a transformer layer. {\color{black}The LCG learns local features within patches, but it ignores global relationships across patches. The transformer conducts multi-head self-attention to learn global relationships across patches, however, it underrates the local feature learning within patches. Thus, the interaction between LCG and transformer is of great significance to make the advantages of each complement the disadvantages of the other. To this end, we make a two-way connection design to achieve a good interaction between LCG and transformer, i.e., one current LCG is embedded after the former level’s LCG and transformer layer, and the current transform is embedded after the current LCG and the former level’s transformer layer.} More detail are described as follows.

\subsection{Local Correlation Graph Module} \label{sec:lcg}

We propose a local correlation graph (LCG) module to aggregate and learn discriminative local features within every patch.
Assume each vehicle image have $N$ patches and the i-th patch {\color{black} $N_{i} \in \Re ^ {P^2 \times C} $ }, where $P^2$ and $C$ respectively denote the patch's size and the number of channels.
In the LCG module, each patch is sampled into $n$ local features of a sequence of $(S, S)$ size with $d$ dimensions features, where $n = \frac{P^2}{S^2}$ and $d = S^2 \times C$.
Like patch embedding of ViT, we flatten $n$ local features $d$ dimensions and map to $d'$ dimensions with a trainable linear projection in every patch.
{\color {black} The transformed local feature sequence of i-th patch $X_i$ is formed according to Eq. \eqref{eq:seq} as follows: }

\begin{equation}\label{eq:seq}
{X}_{i}=\left[{x}_{i}^{1} {E} , {x}_{i}^{2} {E} , \cdots , {x}_{i}^{m} {E} , \cdots , {x}_{i}^{n} {E}\right],
\end{equation}
where ${x}_{i}^{m}  \in \mathbb{R}^{d} $ is the m-th local feature of the i-th patch, ${X}_{i} \in \mathbb{R}^{n \times d'}$, and $\quad {E} \in \mathbb{R}^{d \times d'}$.
Then, we create a learnable position embedding $E_{pos}$ to preserve the position information for each local feature, where ${E}_{p o s} \in \mathbb{R}^{n \times d'}$ .
Therefore, according to Eq. \eqref{eq:pos}, the initial embedding sequence of i-th patch is constructed as follows:
\begin{equation}\label{eq:pos}
{X}_{i}^{init}=\left[{x}_{i}^{1} {E} , {x}_{i}^{2} {E} , \cdots , {x}_{i}^{m} {E}, \cdots , {x}_{i}^{n} {E}\right]+{E}_{p o s} .
\end{equation}

For ease of description, the embedding sequence definition is assumed to be identical on each patch.
Specifically, we define the node with $d'$ dimensions features as $V= [{v}_{1} , {v}_{2} , \cdots ,{v}_{m} , \cdots , {v}_{n}]\in \mathbb{R}^{n \times d'}$, where the $m$-th node $v_{m} = {x}^{m} {E} + {E}_{p o s}^{m} \in \mathbb{R}^{d'} $ and $m \in[ 1,2, \cdots, n]$ .
Then, a spatial graph of every patch can be described as  $G = \{{V}\in \mathbb{R}^{n \times d'}, E\in \mathbb{R}^{n \times n}\}$.
The spatial graph’s edges are constructed according to Eq. \eqref{eq:edge}, as follows:
\begin{equation}\label{eq:edge}
E_{v_{i, j}}=\frac{\exp \left(F_{\color{black}Cosine}\left(v_{i}, v_{j}\right)\right)}{\sum_{k=1}^{n} \exp \left(F_{\color{black}Cosine}\left(v_{i}, v_{k}\right)\right)},
\end{equation}
where $i, j\in[ 1,2, \cdots, n]$; $E_{v_{i, j}}$ is the edge between the node $v_i$ and the node $v_j$ in a patch.
$F_{\color{black}Cosine}(v_{i}, v_{j})$ means similarity score of the node $v_i$ and the node $v_j$.
Moreover, it has been empirically shown that the score of the cosine distance  $F_{\color{black}Cosine}$ is more effective in computing the correlation.
And the cosine distance  $F_{\color{black}Cosine}$ is calculated according to Eq. \eqref{eq:cos}, as follows:
\begin{equation}\label{eq:cos}
F_{{\color{black}Cosine}}\left(v_{i}, v_{j}\right)=\frac{v_{i} \cdot v_{j}}{||v_{i}|| ||v_{j}||},
\end{equation}
where $\cdot$ and $||\cdot ||$ respectively denote element-wise multiplication operation and L2 regularization.

From Eq. \eqref{eq:pos}, Eq. \eqref{eq:edge} and Eq. \eqref{eq:cos}, we can easily obtain the graph's adjacency matrix $A_{i,j} = E_{v_{i, j}}\in \mathbb{R}^{n \times n}$ in a patch.
To effectively mine and learn the relationship between discriminative local features, we adopt the graph network to aggregate and update nodes by information propagation from each node to its neighbors' nodes in the graph.
And the aggregation node $U$ of $i$-th graph is updated according to Eq. \eqref{eq:agg}, as follows:
\begin{equation}\label{eq:agg}
U = (D^{-\frac{1}{2}}AD^{-\frac{1}{2}} X_i) \cdot W ,
\end{equation}
where $A \in \mathbb{R}^{n \times n}$ is the adjacent matrix and $D\in \mathbb{R}^{n \times n}$ is the degree matrix of $A$ ;
The $\cdot$ and $W\in \mathbb{R}^{n \times d'}$  represent an element-wise multiplication operation and a learnable parameter, respectively.
It is worth noted that we use standard 2D learnable parameters $W$ since we have not observed significant performance gains from using more advanced 3D learnable parameters $W$.
In other words, $W\in \mathbb{R}^{n \times d'}$ is shared for $N$ graphs of a patch, which significantly reduces parameter calculation complexity.

From Eq. \eqref{eq:agg}, both a node itself and its neighbor nodes are aggregated and updated according to the learnable weight $W$ to feed more local information.
To introduce non-linearities and improve the convergence graph network, the node $U$ is enhanced according to Eq. \eqref{eq:agg}, as follows:
\begin{equation} \label{eq:act}
O = GELU(LN(U)),
\end{equation}
where $GELU$  represents the gaussian error linerar units (GELU) and LN denotes the layer normalization (LN);
There is nothing special about choosing GELU and LN functions, just to be consistent with the original transformer layer's functions.
According to the Eq. \eqref{eq:act}, $O \in \mathbb{R}^{n \times d'}$ replaces $U$ as the new node with discriminative local features of a graph.

{\color{black}

%\subsection{Pyramidal Graph Network (PGN)}
\subsection{Transformer Layer} \label{sec:tl}
The transformer layer is used to model the global relationship between the different patches, consisting of a multi-head self-attention (MHSA) and a multi-layer perceptron (MLP) block.
Assume that $X \in \mathbb{R}^{N \times M}$ is $N$ patches, where $M = P^2 \times C$.
The $(P, P)$ is the size of each image patch and $C$ is the number of channels.
The $X \in \mathbb{R}^{N \times M}$  are linearly transformed to queries $Q \in \mathbb{R}^{N \times M_{q}}$ ,
keys $K \in \mathbb{R}^{N \times M_{k}}$  and values $V \in \mathbb{R}^{N \times D_{v}}$.
The scaled dot-product self-attention is applied on $Q$, $K$, $V$ according to the Eq. \eqref{eq:attn}, as follows:
\begin{equation}\label{eq:attn}
\text { Attention }(Q, K, V)=\operatorname{softmax}\left(\frac{Q K^{T}}{\sqrt{M_{v}}}\right) V.
\end{equation}
The MHSA splits the queries, keys, and values for $h$ times and performs the $h$ times self-attention function in parallel.
Then the output values of each head are concatenated and linearly projected to form the final output.
Therefore, according to Eq. \eqref{eq:msa} and Eq. \eqref{eq:mlp}, The output of transformer layer $Y\in \mathbb{R}^{N \times M}$ is calculated as follows:
\begin{equation}\label{eq:msa}
X'= MSA(LN(X)) + X ,
\end{equation}
\begin{equation}\label{eq:mlp}
Y = MLP(LN(X') + X' .
\end{equation}
From Eq. \eqref{eq:attn}, Eq. \eqref{eq:msa} and Eq. \eqref{eq:mlp}, one can see that since the transformer layer models global features on all patches, the proposed LCG module is essential for local feature learning.

\begin{figure}[tp]
    \centering
    \includegraphics[width=1\linewidth]{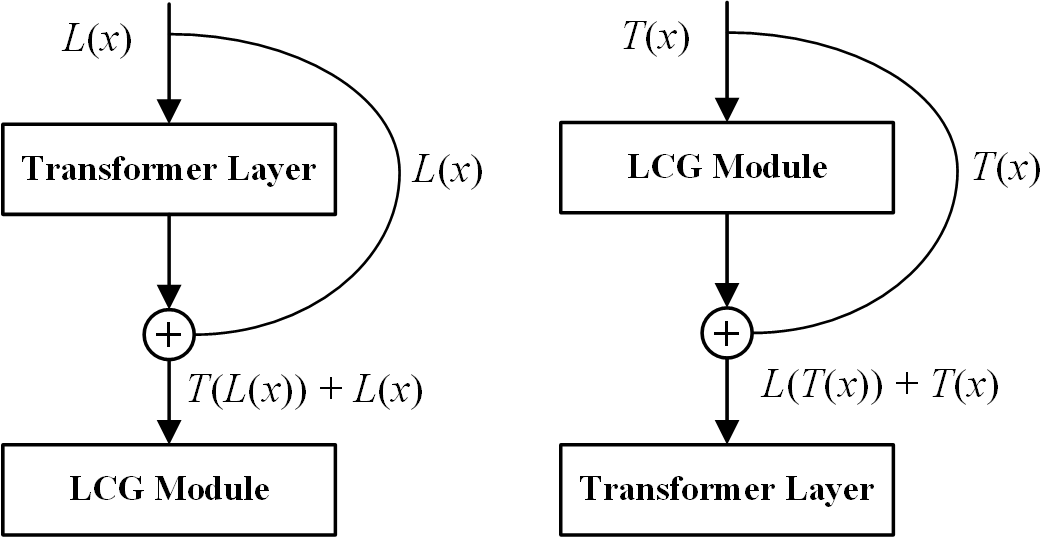}
    \vspace{-.35cm}
    \caption{The transformer layer and the LCG module interact each other.}

\label{fig:couple}
\end{figure}

\subsection{Graph Interactive Transformer }
{
	\color{black}
As shown in Fig. \ref{fig:framework} , the graph interactive transformer (GiT) method has $l$ GiT blocks.
Every GiT block consists of a local correlation graph (LCG) module and a transformer layer.
It is worth noting that our proposed GiT block can make the global features of the transformer layer and the local features of the LCG module form a coupling state.
In other words, global features and local features can interact with each other in the entire feature learning process, as shown in Fig. \ref{fig:couple}.
Specifically, as describe in section \ref{sec:lcg}, assume that $O^{l} \in \mathbb{R}^{N \times n \times d'}$ represents the LCG module of the $l$-th GiT block has $N$ graph with discriminatory local feature.
And each graph is composed of $n$ nodes with $d'$ dimensional features.
When embedding local features in the patch,
{\color{black} we set $d' = d$ to make a LCG and a transformer layer can be directly coupled without requiring extra parameterized modules, such as linear projection or convolution layers, saving parameters and computational cost.}
According to  Eq. \eqref{eq:reshape}, the discriminative local feature $O^{l}$ can be seamlessly converted into feature $X^{l}$ of a patch.
\begin{equation}\label{eq:reshape}
X^{l} = Reshape(O^{l})
\end{equation}
where the $O^{l} \in \mathbb{R}^{N \times m \times d'}$ and the $X^{l} \in \mathbb{R}^{N \times D}$.
From node to patch, the $O^{l}$ can be reshape $N$ patches with D dimensions features and $l \in[ 1,2, \cdots, 12] $ in this paper.

Let’s forget about specific formulas of section \ref{sec:lcg} and section \ref{sec:tl}, the $l$-th GiT blocks' LCG module $L^{(l)}$ and transformer layer $T^{(l)}$  are defined according to Eq. \eqref{eq:lgc} and Eq. \eqref{eq:tl}, as follows:
\begin{equation}\label{eq:lgc}
{\color{black} L^{(l)}=L(T^{(l-1)}(L^{(l-1)}(x)), L^{(l-1)}(x)), l\ge 2 ,}
\end{equation}
\begin{equation}\label{eq:tl}
{\color{black}  T^{(l)}=T( L^{(l)}(T^{(l-1)}(x)), T^{(l-1)}(x)), l\ge 2 .}
\end{equation}

From the  Eq. \eqref{eq:lgc} and Eq. \eqref{eq:tl}, in the $l$-th GiT block, the LCG module learns local features from local and global features resulting from the LCG module and transformer layer of the $(l-1)$-th GiT block.
Similarly, the transformer layer learns global features from the global relationship generated by the transformer layer of the $(l-1)$-th GiT block and the new local features outputted via the LCG module of the $l$-th GiT block.
Therefore, GiT blocks' LCG modules and transformer layers are in a coupled status, bringing effective cooperation between local and global features.
}

\begin{table}[tp]
	\renewcommand{\arraystretch}{1.0}
	%\vspace{-.2cm}
	\caption{The node configuration for LCG modules.
		%Here, $h$ and $w$ denote the height and width, respectively.
	} \label{tab:config}
	%\small
\vspace{-.2cm}
	\begin{center}
		\begin{tabular}{c|ccc}
			\hline
			  &
			\begin{tabular}{c}
				Sampling Size\\
				($S \times S $)\\
			\end{tabular}
			&
			\begin{tabular}{c}
 Number \\
($m$)\\
	\end{tabular}
&
\begin{tabular}{c}
Dimension\\
($d$)\\
	\end{tabular}
			%& $d'$
\\
\hline
			Stage 1              & $2 \times 2$       & 64                                            & 12 \\%& 12\\
			Stage 2             & $4 \times 4$        & 16                                             & 48 \\%& 48    \\
			Stage 3             & $8 \times 8$   & 4                                                    & 192 \\%& 192    \\
			\hline
		\end{tabular}
	\end{center}
%\vspace{-.3cm}
\end{table}

Besides, the GiT method has three stages and every stage has 4 GiT blocks to handle feature maps of a specific scale.
Each GiT block has the same architecture.
It's just that the number of nodes $n$ and the sampling size $(S, S)$ are different in different stages.
In the subsequent experiments, each image patch's resolution is the same as the ViT \cite{vit} method, i.e., the size of each patch is $(16, 16)$.
The specific configuration of the node size in the LCG module is shown in Table \ref{tab:config}.

{\color{black}
In addition to the two-directional way in Eq.\eqref{eq:lgc} and Eq. \eqref{eq:tl}, there are two other skip-connection architectures, namely, GiT (Global $\rightarrow$ Local) and GiT (Global $\leftarrow$ Local), which have a single-directional feature flow. The GiT (Global $\rightarrow$ Local) case means that it only uses blue connection line in Fig. \ref{fig:framework}. The GiT (Global $\leftarrow$ Local) case means that it only uses  red connection line in Fig. \ref{fig:framework}.  Intuitively, two-directional way could couple global and local features more completely than the GiT (Global $\rightarrow$ Local) and GiT (Global $\leftarrow$ Local).
}
\subsection{Loss Function Design}
{\color{black}
From Fig. \ref{fig:framework}, the class token of the last GiT block serves as the image's vehicle feature representation.
Inspired by BNNeck \cite{luo2019strong} in the CNN, we introduce it after the final class token.
The features of the class token are directly applied to triplet loss of soft margin.
After the feature of batch normalization (BN) is fed to cross-entropy loss without label smoothing.
Therefore, the proposed GiT's total loss function is formulated as follows:

\begin{equation}\label{eq:totoal}
%\color{blue}
{L_{total}} = \alpha L_{CE} + \beta L_{Triplet},
\end{equation}

where $L_{CE}$ and {\color{black}$L_{Triplet}$}  respectively denote cross-entropy loss and  triplet loss.
$\alpha$ and $\beta$ are manually setting constants used to keep balance of two loss functions.
To avoid excessive tuning those constants, we set $\alpha= \beta = 1$ in following experiments.
The cross-entropy loss function \cite{LSR} is formulated as follows:
\begin{equation}\label{eq:llsrs}
{%\color{blue}
    {L_{CE}}(X,l) = \frac{{ - 1}}{M}\sum\limits_{i = 1}^M {\sum\limits_{j = 1}^K { ({l_i},j)log(\frac{{{e^{W_j^{\rm{T}}{X_i}}}}}{{\sum\nolimits_{k = 1}^K {{e^{W_k^{\rm{T}}{X_i}}}} }})} },
}
\end{equation}

where $X$ is a training set and $l$ is the class label information;
$M$ is number of training samples;
$K$ is the number of classes;
$(X_i, l_i)$ is the $i$-th training sample and $l_i \in \{1, 2, 3, ..., K\}$; $W=[W_1, W_2, W_3, ..., W_K]$ is a learn-able parameter matrix.

The triplet loss function \cite{triplet} is formulated as follows:
\begin{footnotesize}
{\begin{equation}\label{eq:triplet}
{\color{black}
{L_{Triplet}}({X^a},{X^n},{X^p}) = \frac{{ - 1}}{M}\sum\limits_{i = 1}^M {\max ({{\left\|{X_i^a - X_i^n} \right\|}_2}-{{\left\|  {X_i^a - X_i^p} \right\|}_2} ,0)},
}
\end{equation}}
\end{footnotesize}
where $({X^a},{X^n},{X^p})$ is a set of training triples; $M$ is number of training triplets; for the $i$-th training triplet, $(X^a_i, X^n_i)$ is a negative pair holding different class labels, and $(X^a_i, X^p_i)$ is a positive pair having the same class label;
$\left\|\cdot\right\|_2$ denotes the an Euclidean distance. Moreover, the hard sample mining \cite{triplet} strategy is applied to improve the triplet loss, which aims to find the most difficult positive and negative image pairs in each mini-batch.
Recently, {\color{black} there is a weighted }regularization triplet (WRT) \cite{wrt} loss function can well constrain the pairwise
distance of a positive/negative sample pair without introducing any extra hyper-parameters for the re-identification task.
Considering that most state-of-the-art methods do not apply WRT loss, our default configuration still applies initial triplet loss $L_{Triplet}$ .

\section{Experiment and Analysis}\label{sec:exp}
To validate the proposed GiT method's superiority, it is compared with multiple state-of-the-art vehicle re-identification approaches on three large-scale datasets, namely, VeRi776 \cite{veri776}, VehicleID \cite{drdl} and VeRi-Wild \cite{wild}.
The rank-1 identification rate (R1) \cite{veri776, drdl, benchmark} and mean average precision (mAP) \cite{market, dnd, qddlf, dnffn} are used to assess the accuracy performance.

\subsection{Datasets}
\textbf{\emph{VeRi776}} \cite{veri776} is captured by 20 cameras in unconstrained traffic scenarios, and each vehicle is captured by 2-18 cameras.
Following the evaluation protocol of \cite{veri776}, VeRi776 is divided into a training subset containing 37,746 images of 576 subjects and a testing subset including a probe subset of 1,678 images of 200 subjects and a gallery subset of 11,579 images of the same 200 subjects.
Moreover, only cross-camera vehicle pairs are evaluated. Suppose the same camera captures a probe image and a gallery image.
In that case, the corresponding result will be excluded from the evaluation process.

\textbf{\emph{VehicleID}} \cite{drdl} includes 221,763 images of 26,267 subjects. Each vehicle is captured from either front or rear viewpoint. The training subset consists of 110,178 images of $13,164$ subjects. There are three testing subsets, i.e., Test800, Test1600 and Test2400, for evaluating the performance at different data scales.
Specifically, Test800 includes 800 gallery images and 6,532 probe images of 800 subjects. Test1600 contains 1,600 gallery images and 11,395 probe images of 1,600 subjects. Test2400 is composed of 2,400 gallery images and 17,638 probe images of 2,400 subjects. {\color{black}Moreover,} for three testing subsets, the division of probe and gallery subsets is implemented as follows: randomly selecting one image of a subject to form the probe subset. All remaining images of this subject are used to construct the gallery subset. This division is repeated and evaluated 10 times, and the average result is reported as the final performance.

{\color{black}
\textbf{\emph{VeRi-Wild}}~\cite{wild} is newly dataset released in CVPR 2019.
Different to the VeRi776 \cite{veri776} and VehicleID \cite{drdl} captured at day, its images captured at both day and night.
VeRi-Wild contains $416,314$ images of $40,671$ subjects in total. It is divided into a training subset of $277,797$ images of $30,671$, and a testing subset of $128,517$ images of $10,000$ subjects. Similar to VehicleID \cite{drdl}, the testing subset is organized into three different scale subsets, i.e., Test3000, Test5000, and Test10000. To be more specific, Test3000 is composed of  $41,816$ gallery images and $3000$ probe images of $3,000$ subjects.
Test5000 is made up of  $69,389$ gallery images and $5,000$ probe images of $5,000$ subjects.
Test10000 is consisted of $138,517$ gallery images and $10,000$ probe images of $10,000$ subjects.
}

\subsection{Implementation Details}
The deep learning toolbox is PyTorch \cite{pytorch} with FP16 training.
Training configurations are summarized as follows.
(1) We respectively use ImageNet pre-trained weights of ViT and a truncated normal distribution initialize to the transformer layers and the LCG modules.
(2) Random erasing \cite{RS} and z-score normalization are used for the data augmentation. Both probabilities of horizontal flip and random erasing are set to 0.5.
(3) The mini-batch stochastic gradient descent (SGD) method \cite{alexnet} is applied to train parameters. The weight decays are set to 1$\times$10$^{-4}$ and the momentums are set to 0.9. There are 150 epochs for the training process. The learning rates are initialized to 1$\times$10$^{-4}$, and they are linearly warmed up \cite{luo2019strong} to 1$\times$10$^{-2}$ in the first 5 epochs. After warming up, the learning rates are maintained at 1$\times$10$^{-2}$ from 11th to 50th epochs. Then, the learning rates are reduced to 1$\times$10$^{-3}$ from {\color{black} 51st} to 85th epochs, {\color{black} it decays to 1$\times$10$^{-4}$ after 85 epochs and further dropped to 1$\times$10$^{-5}$ between 120th and 150th epochs.}
(4) Unless otherwise specified, each mini-batch includes 8 subjects and each subject holds 6 images.

During the testing phase, those 768-dimensional features resulted from the BN layer (see Fig. \ref{fig:framework}) as the vehicle images' final features.
Moreover, the Cosine distance of those the final features is applied as the similarity measurement for vehicle re-identification.

%Euclidean

\subsection{Comparison with State-of-the-art Methods}
According to the development process of vehicle re-identification, those state-of-the-art methods under comparison are roughly classified into three categorizes for a clear presentation.
Specifically, Global (Pure CNN) denotes the approaches \cite{sdccnn, vst, veri2017, dense, oife, veri2016, largecar, drdl, djdl, pmsm, mattribute} exploiting global features based on pure convolutional neural network (CNN);
Global + Local (Pure CNN) represents the methods \cite{saver, vocreid, pven, app, cfvmnet, sff, partreg, san, pamtri, mrm, dmml, vanet, aaver, qddlf, vami} utilizing global features and local features  based on pure CNN;
Global + Local (CNN $\&$ GN) denotes the approaches \cite{liu_gcn, zhu2020graph, hpgn} dealing with global features and local features based on CNN and graph network (GN).

\begin{table}[tp]
    \renewcommand{\arraystretch}{1.0}
    \caption{The performance (\%) comparison on VeRi776. The {\color{red}{red}}, {\color{green}{green}} and {\color{blue}{blue}} rows represent the {\color{red}{$1$st}}, {\color{green}{$2$nd}} and {\color{blue}{$3$rd}} places, respectively.
    } \label{tab:veri776}
    \vspace{-.2cm}
    \begin{center}
        %	\small
        %\setlength{\tabcolsep}{6pt}
        \begin{threeparttable}
            \begin{tabular}{c|c|cc}
                \hline
                \multicolumn{2}{c|} {Methods}  &R1  & mAP   \\
                \hline
                %\noalign{\smallskip}
                \multirow{1}{*}{{Proposed}}
                &{GiT}
                &{\color{red}{96.86}}
                &{\color{red}{80.34}}\\
                \hline
                \multirow{1}{*}{{Transformer}}
                &{*ViT} \cite{vit}
                &{95.84}
                &{78.92}

                \\ %E-element
                \hline
                \multirow{2}{*}{\begin{tabular}[c]{@{}c@{}}Global + Local\\ (CNN \& GN)\end{tabular}}

                %&{HPGN \cite{hpgn}} & \color{blue}{96.72} & \color{green}{80.18} \\
                &*PCRNet \cite{liu_gcn} & 95.40 &78.60  \\
                &*SGAT \cite{zhu2020graph} & 94.65 &76.32 \\
                %&HSS-GCN \cite{hssgcn} &64.40  &44.80 \\
                \hline
                \multirow{15}{*}{\begin{tabular}[c]{@{}c@{}}Global + Local\\ (Pure CNN)\end{tabular}}
                %			BSVR \cite{bsvr} &90.46 &70.74 &ICMM 2020\\
                &*SAVER \cite{saver}           &\color{green}{96.40}&\color{blue}79.60 \\
                &*VOC-ReID \cite{vocreid}           &\color{blue}{96.30}&\color{green}79.70\\
                &*PVEN \cite{pven}           &95.60&79.50 \\
                %PGAN \cite{pgan}                                                            &96.5 & 79.3 & arXiv\\
                &{App+License \cite{app}  } &{95.41} &{78.08} \\
                &*SFF+SAtt \cite{sff} & {95.35} & {77.28} \\
                &{CFVMNet \cite{cfvmnet}  } &{95.30} &{77.06} \\
                &Part Regular \cite{partreg}&94.30&74.30   \\
                &*SAN \cite{san} &94.82 &74.68   \\
                &PAMTRI \cite{pamtri} & 92.86 & 71.88  \\
                &MRM \cite{mrm} &91.77 &68.55   \\
                %	CCA \cite{cca}  &91.71 & 68.05   & arXiv 2019\\
                &DMML \cite{dmml} &91.20 &70.10  \\
                &VANet \cite{vanet}                  &89.78&66.34   \\
                &AAVER \cite{aaver} &88.97 &61.18  \\
                &QD-DLF \cite{qddlf}                                                     &88.50&61.83\\
                &VAMI \cite{vami}                                                             &85.92&61.32  \\
                \hline
                \multirow{6}{*}{\begin{tabular}[c]{@{}c@{}}Global\\ (Pure CNN)\end{tabular}}

                %&RAM \cite{ram}                                                                   &88.60&61.50  \\
                &SDC-CNN \cite{sdccnn}                                                  &83.49&53.45  \\
                &VST Path Proposals \cite{vst}                                      &83.49&58.27  \\ %
                &PROVID \cite{veri2017}                                                     &81.56&53.42  \\
                &*DenseNet121 \cite{dense}                                          &82.77&54.83   \\
                %&NuFACT \cite{veri2017}                                                     &76.76&48.47  \\
                &*GoogLeNet \cite{largecar}                                             &78.64&51.42  \\
                &OIFE+ST \cite{oife}                                                         &68.30&51.42  \\
                &  FACT \cite{veri2016}                                                    &52.21&18.75   \\

                \hline
            \end{tabular}
            \begin{tablenotes}
                \footnotesize
                {\color{black}\item[*] Methods the same training tricks as GiT.}
            \end{tablenotes}
        \end{threeparttable}
    \end{center}
    %\vspace{-0.4cm}
\end{table}
%\vspace{-.2cm}

\begin{table}[tp]
    \renewcommand{\arraystretch}{1.0}
    \caption{The performance (\%) comparison on VehicleID. The {\color{red}{red}}, {\color{green}{green}} and {\color{blue}{blue}} rows represent the {\color{red}{$1$st}}, {\color{green}{$2$nd}} and {\color{blue}{$3$rd}} places, respectively.
    }\label{tab:vehicleid}
    \vspace{-.2cm}
    \begin{center}
        %\normalsize
        \footnotesize
        \setlength{\tabcolsep}{1. pt}
        \begin{tabular}{c|c|cc|cc|cc}
            \hline
            \multicolumn{2}{c|}{\multirow{2}{*}{Methods}}
            &\multicolumn{2}{c|} {Test800}
            & \multicolumn{2}{c|} {Test1600}
            & \multicolumn{2}{c}{Test2400}
            \\
            %            \cline{2-9}
            \multicolumn{2}{c|}{}
            & R1     &mAP
            & R1      &mAP
            %& R1      &mAP
            & R1      &mAP \\

            %\noalign{\smallskip}
            \hline
            \multirow{1}{*}{{Proposed}}
            &{GiT}
            &\color{red}{84.65}&\color{red}{90.12}
            &\color{red}{80.52}&\color{red}{86.77}
            &\color{red}{77.94}&\color{red}{84.26}

            \\
            \hline

            \multirow{1}{*}{{Transformer}}
            &{ViT} \cite{vit}
            &\color{blue}{82.41}&\color{blue}{87.05}
            &\color{blue}{76.36}&\color{blue}{83.29}
            &\color{blue}{73.31}&\color{blue}{80.10}
            %&\color{red}{80.40}&\color{red}{86.45}
            \\

            \hline

            \multirow{3}{*}{\begin{tabular}[c]{@{}c@{}}Global + Local\\ (CNN \& GN)\end{tabular}}

            &HPGN \cite{hpgn}	&\color{green}{83.91}&\color{green}{89.60}&\color{green}{79.97}&\color{green}{86.16}&\color{green}{77.32}&\color{green}{83.60} \\
            &SGAT \cite{zhu2020graph} & 78.12&81.49 &73.98 & 77.46 &71.87 &75.35 \\
            &HSS-GCN \cite{hssgcn} &72.70 &77.30 & 67.90 &72.40 &62.40  &66.10 \\

            \hline
            \multirow{5}{*}{\begin{tabular}[c]{@{}c@{}}Global + Local\\ (Pure CNN)\end{tabular}}

            &{App+License \cite{app}}
            &{79.50} &{82.70}
            &{76.90} &{79.90}
            &{74.80} &{77.70}
            %&\color{green}{77.06} &\color{green}{80.1}
            \\

            &{MGL \cite{mgl}}
            &{79.60} &{82.10}
            &{76.20} &{79.60}
            &{73.00} &{75.50}
            %&\color{blue}{76.26} &\color{blue}{79.06}
            \\

            &MSV \cite{msv}
            &75.10 &79.30
            &71.80 &75.40
            &68.70 &73.30
            %&71.86 &76
            \\

            &EALN \cite{ealn}
            &75.11 &77.50
            &71.78 &74.20
            &69.30 &71.00
            %&71.06 &74.23
            \\

            &QD-DLF \cite{qddlf}
            &{72.32} &{76.54}
            &{70.66}&{74.63}
            &{64.14} &{68.41}
            %&{69.04} &{73.19}
            \\

            \hline
            \multirow{2}{*}{\begin{tabular}[c]{@{}c@{}}Global\\ (Pure CNN)\end{tabular}}

            &SDC-CNN \cite{sdccnn}
            &{56.98} &63.52
            &{50.57}   &57.07
            &{42.92} &49.68
            %&{50.15} &56.75
            \\
            &DenseNet121 \cite{dense}
            &{66.10}  &68.85
            &{67.39}      &{69.45}
            &{63.07}   &{65.37}
            %&{65.53}   &{67.89}

            \\
            \hline
        \end{tabular}

    \end{center}
    %\vspace{-.4cm}
    %\vspace{-.6cm}
\end{table}

\begin{table}[tp]
	\renewcommand{\arraystretch}{1.0}
	\caption{The performance (\%) comparison on the VeRi-Wild. The {\color{red}{red}}, {\color{green}{green}} and {\color{blue}{blue}} rows represent the {\color{red}{$1$st}}, {\color{green}{$2$nd}} and {\color{blue}{$3$rd}} places, respectively.
	}\label{tab:wild}
	%\vspace{-.3cm}
	\begin{center}
		\footnotesize
		\setlength{\tabcolsep}{2pt}
		\begin{tabular}{c|c|cc|cc|cc}
			\hline
			%\noalign{\smallskip}
\multicolumn{2}{c|}{\multirow{2}{*}{Methods}}

			&\multicolumn{2}{c} {Test3000}
			&\multicolumn{2}{c} {Test5000}
			&\multicolumn{2}{c}{Test10000}
			\\
\multicolumn{2}{c|}{}
			%\cline{2-7}
			& R1     &mAP
			& R1      &mAP
			& R1      &mAP \\
			%\noalign{\smallskip}

			\hline
\multirow{1}{*}{{Proposed}}
			&{GiT}
			&\color{red}{92.65}&\color{red}{81.76}
			&\color{red}{89.92}&\color{red}{75.64}
			&\color{red}{85.41}&\color{red}{67.55}
			%&\color{red}{80.40}&\color{red}{86.45}
			\\
\hline
\multirow{1}{*}{{Transformer}}
			&{ViT} \cite{vit}
			&{89.29}&{78.66}
			&{86.18}&{72.73}
			&{81.13}&{63.91}
			%&\color{red}{80.40}&\color{red}{86.45}
			\\
\hline
\multirow{2}{*}{\begin{tabular}[c]{@{}c@{}}Global + Local\\ (CNN \& GN)\end{tabular}}
&PCRNet \cite{liu_gcn} &\color{green} {92.50} & \color{green} {81.20} &\color{green} {89.60} &\color{green} {75.30} &\color{green} {85.00} & \color{green} {67.10}  \\

			&{HPGN \cite{hpgn}}
			&{91.37}&\color{blue} {80.42}
			&\color{blue}{88.21}&\color{blue}{75.17}
			&\color{blue}{82.68}&\color{blue}{65.04}
			\\

\hline
\multirow{4}{*}{\begin{tabular}[c]{@{}c@{}}Global + Local\\ (Pure CNN)\end{tabular}}
&GLAMOR	\cite{glamor} &\color{blue} {92.10} & 77.15 &N/A &N/A &N/A &N/A  \\
           &UMTS \cite{umts}
			&84.50 &72.70
			&79.30 &66.10
			&72.80 &54.20	\\

            &DFLNet \cite{dflnet}
			&80.68 &68.21
			&70.67 &60.07
			&61.60 &49.02	\\

			&AAVER \cite{aaver}
			&75.80 &62.23
			&68.24 &53.66
			&58.69 &41.68	\\

\hline
\multirow{7}{*}{\begin{tabular}[c]{@{}c@{}}Global\\ (Pure CNN)\end{tabular}}

			&FDA-Net \cite{wild}
			&64.03 &35.11
			&57.82 &29.80
			&49.43 &22.78
			\\
			&{GSTE} \cite{gste}
			&{60.46} &{31.42}
			&{52.12} &{26.18}
			&{45.36} &{19.50}
			\\
			&GoogLeNet \cite{googlenet}
			&57.16 &24.27
			&53.16 &24.15
			&44.61 &21.53
			\\
			&HDC \cite{hdc}
			&57.10 &29.14
			&49.64 &24.76
			&43.97 &18.30
			\\
			&DRDL \cite{drdl}
			&56.96 &22.50
			&51.92 &19.28
			&44.60 &14.81
			\\
			&Softmax \cite{veri776}
			&53.40 &26.41
			&46.16 &22.66
			&37.94 &17.62
			\\
			&Triplet \cite{facenet}
			&44.67 &15.69
			&40.34 &13.34
			&33.46 &9.93
			\\
			\hline
		\end{tabular}
	\end{center}
	%\vspace{-.3cm}
\end{table}

\subsubsection{Comparisons on VeRi776}

 Table \ref{tab:veri776} shows the performance comparison of the proposed GiT method and multiple state-of-the-art approaches on VeRi776 dataset.
It can be found that the proposed GiT method achieves the highest rank-1 (R1) identification rate (i.e., 96.86\%) and mAP (i.e., 80.34\%).
 As shown in Table \ref{tab:veri776}, the method of combining global and local features (i.e., Global + Local (Pure CNN) and Global + Local (CNN $\&$ GN)) are mostly superior to the method of only using global features (Global (Pure CNN)) on the VeRi776 dataset by a large margin.
 For example, the best Global + Local (CNN $\&$ GN) method (i.e., PCRNet \cite{liu_gcn}) and the best Global + Local (Pure CNN) method (i.e., SAVER \cite{saver}) respectively defeats the best Global (Pure CNN) method (i.e.,SDC-CNN \cite{sdccnn}) by 11.91 \% and 12.91\%  in term of R1 identification rate.
 This demonstrates that part details and local features are important clues for vehicle re-identification.
 Meanwhile,  the R1 identification rate and mAP of the GiT approach exceeds 13.37\% and 26.89 \% over the best Global (Pure CNN) method (i.e.,SDC-CNN \cite{sdccnn}).

Moreover, among those compared state-of-the-art methods, the proposed GiT approach outperforms the best Global + Local (CNN $\&$ GN) method (i.e., PCRNet \cite{liu_gcn}) and the best Global + Local (Pure CNN) method.
For example, the proposed GiT method respectively higher 1.46\% and 1.74\% than the best Global + Local(CNN $\&$ GN) method (i.e., PCRNet \cite{liu_gcn}) on  R1 identification rate and mAP. It is noted that the PCRNet \cite{liu_gcn} extra uses an image segmentation model trained on vehicle parsing data as a preprocessing tool to obtain the parsed masks of vehicle images, while GiT does not use any extra semantic annotation information.

\subsubsection{Comparisons on VehicleID}
%More analysis are described as follows.
Table \ref{tab:vehicleid} shows the rank-1 identification rate (R1) and mAP comparison of the proposed GiT method and state-of-the-art approaches on the VehicleID \cite{drdl} dataset with a more significant number of images than the VeRi776 \cite{veri776} dataset.
Three different type of testing subsets are evaluated, including Test800, Test1600, and Test2400.
It can be found that the proposed GiT method consistently outperforms those state-of-the-art methods under comparison.
{\color {black}
For example, on Test800, GiT defeats the second place HPGN \cite{hpgn} by a 0.52\% higher R1 identification
rate, and outperforms the third place ViT \cite{vit} by a 3.07\% higher R1 identification rate.

}

\subsubsection{Comparisons on VeRi-Wild}
Table \ref{tab:wild} shows the comparison result on VeRi-Wild \cite{wild} dataset.
Among those methods, it can be found that the proposed GiT method obtains the 1st place again on the Test3000, Test5000, and Test1000 of VeRi-Wild dataset.
For example, on largest Test1000 subset, R1 and mAP of GiT method respectively are 0.41\% and 0.45\% higher than those of the 2nd place method, i.e., PCRNet \cite{liu_gcn}, which extra uses an image segmentation model.
Compared with the 3rd place method (i.e., HPGN \cite{hpgn}), the proposed GiT method defeats it by 2.73\% in term of mAP and 2.51\% in term
of R1 identification rate on largest Test1000 subset.
Meanwhile, the proposed GiT method obtains the state-of-the-art performance on VeRi776, VehicleID, and VeRi-Wild, which shows the effectiveness and generalization of our method.
Besides, one can see that a lot of approaches (the Global + Local \cite{umts,dflnet,glamor,aaver,liu_gcn,hpgn}) have made significant progress than the Global (Pure CNN) methods \cite{wild, gste, googlenet, hdc, drdl, veri776, facenet} on the  VeRi-Wild dataset.
Those results also reflect the trend that the methods which exploit both global and local features achieve better results than early Global (Pure CNN) methods that only use global features.
}

{\color{black}
As shown in Table \ref{tab:veri776}, Table  \ref{tab:vehicleid}, and Table \ref{tab:wild}, our GiT method consistently outperforms ViT on three datasets. Especially, on two large sized dataset, namely, VehicleID and VeRi-Wild, GiT wins obvious improvements. For example, on Test2400 of VehicleID, our GiT’s R1 is 1.02\% higher than ViT’s R1, and our GiT’s mAP is 4.16\% larger than the ViT’s mAP. These results demonstrate that GiT coupling graphs is effective to improve ViT.
}

\begin{table}[tp]
    \renewcommand{\arraystretch}{1}
    %\vspace{-.2cm}
    \caption{The ablation study of our GiT method on the VeRi776 and VehicleID datasets.}
    \label{tab:ablation}
    %\vspace{-.25cm}
    \begin{center}
        %\scriptsize
        \footnotesize
        \setlength{\tabcolsep}{1.5pt}
        \begin{tabular}{l|cc |cc cc cc }
            \hline
            %\noalign{\smallskip}
            \multirow{3}{*}{Names}
            %			&
            %			\multirow{3}{*}{Scales}
            & \multicolumn{2}{c|}{\multirow{2}{*}{VeRi776}}
            &\multicolumn{6}{c}{VehicleID}
            \\
            & &
            & \multicolumn{2}{c}{Test800}
            & \multicolumn{2}{c}{Test1600}
            & \multicolumn{2}{c}{Test2400}
            \\

            & R1      &mAP
            & R1      &mAP
            & R1      &mAP
            & R1      &mAP
            \\
            %\noalign{\smallskip}
            \hline
            Baseline (Global)  %(Baseline)
            &95.84 &78.92
            &82.41 &87.05
            &76.36 &83.29
            &73.31 &80.10
            \\
            GiT (No Interactive)
            &96.17 &79.55
            &83.26 &88.88
            &79.05 &85.29
            &76.22 &82.50
            \\
            GiT (Global $\rightarrow$ Local)
            &96.34 &79.78
            &83.57 &89.13
            &79.29 &85.52
            &76.46 &82.61
            \\
            GiT (Global $\leftarrow$ Local)
            &96.42 &79.87
            &83.61 &89.16
            &79.43 &85.66
            &76.75 &82.97
            \\
            GiT (Interactive)
            &\textbf{96.86} &\textbf{80.34}
            &\textbf{84.65} &\textbf{90.12}
            &\textbf{80.52} &\textbf{86.77}
            &\textbf{77.94} &\textbf{84.26}
            \\
            \hline
        \end{tabular}
    \end{center}
    %\vspace{-.2cm}
\end{table}

\begin{figure}[tp]
    \centering
    \includegraphics[width=.8\linewidth]{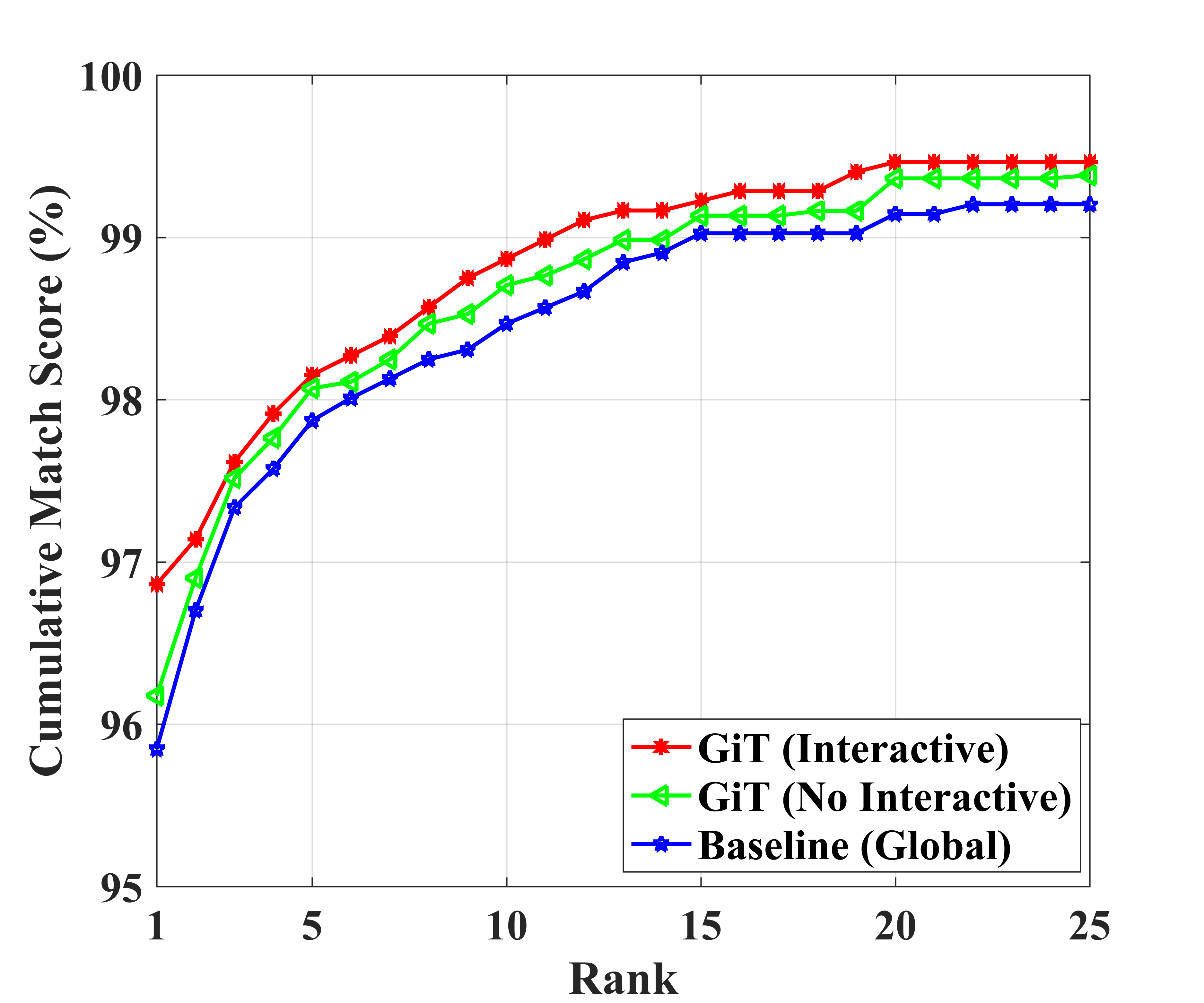}
    %\vspace{-.7cm}
    \caption{The CMC cures on VeRi776 dataset.}
    \label{fig:776}
    %\vspace{-.6cm}
\end{figure}

\begin{figure}[tp]
    \centering
    \includegraphics[width=.8\linewidth]{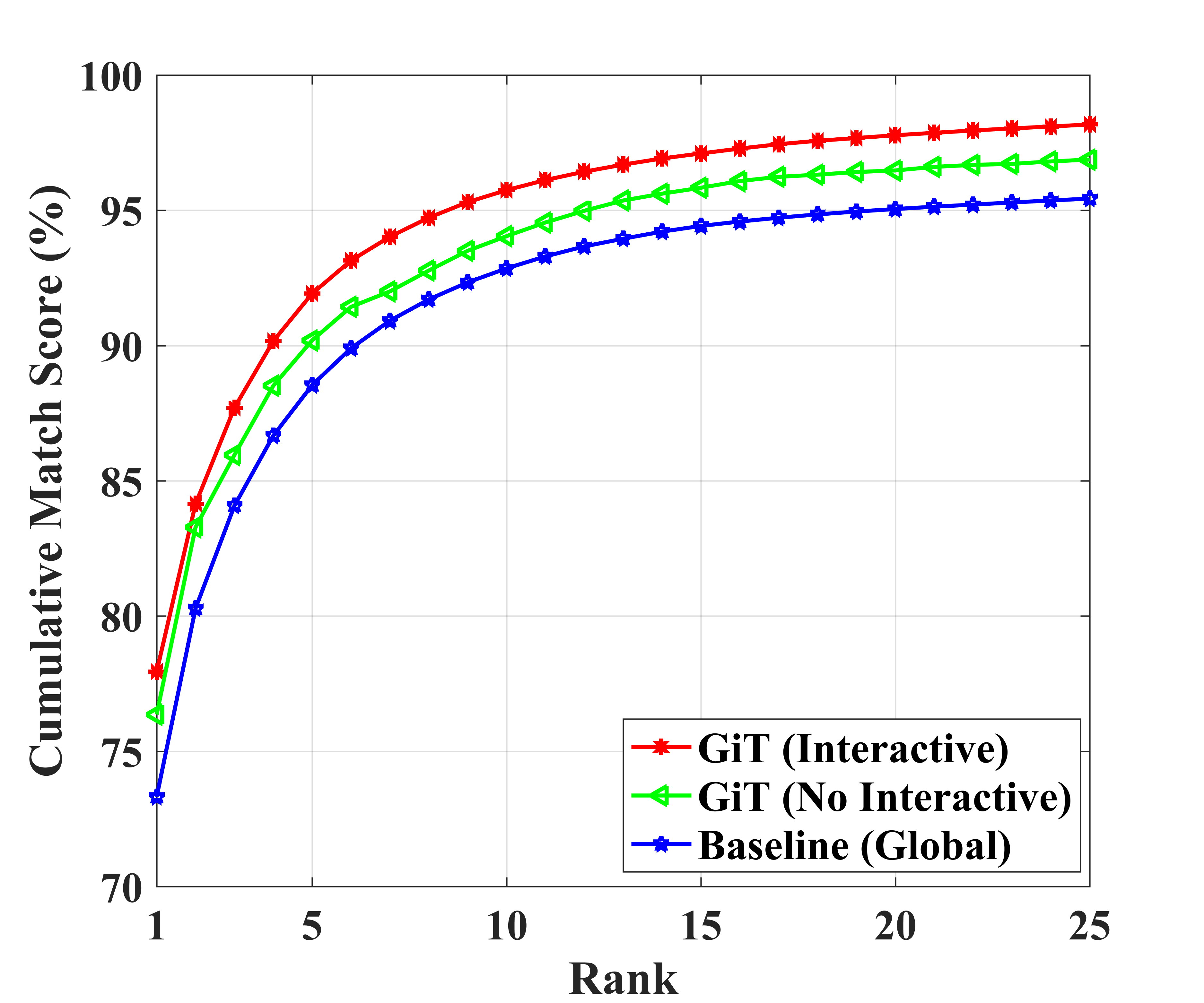}
    %\vspace{-.7cm}
    \caption {The CMC cures on VehicleID dataset.}
    \label{fig:2400}
    %\vspace{-.6cm}
\end{figure}

\subsection{Ablation Studies and Analysis}
The comparison results presented in Table \ref{tab:veri776}, Table  \ref{tab:vehicleid}, and Table \ref{tab:wild}, demonstrate that the proposed GiT method is superior to many state-of-the-art vehicle re-identification methods.
Recall Fig. \ref{fig:framework}, the proposed GiT method is mainly made up of pure transformer layers for learning global features and local correlation graph (LCG) modules for mining local features.
Moreover, the LCG module's local features and global features of the transformer layer can interact and improve each other.
In what follows,  the proposed GiT method is comprehensively analyzed from six aspects to investigate the logic behind its superiority.
{\color{black}(1) Role of LCG module and transformer layer interaction.
(2)  Visualization.
(3) Influence of sampling strategy in LCG module.
(4)  Role of backbones.
(5)   Complexity analysis.
 (6)  Sensitivity analysis.

}

\begin{figure}[tp]
	\centering
	\includegraphics[width=.9\linewidth]{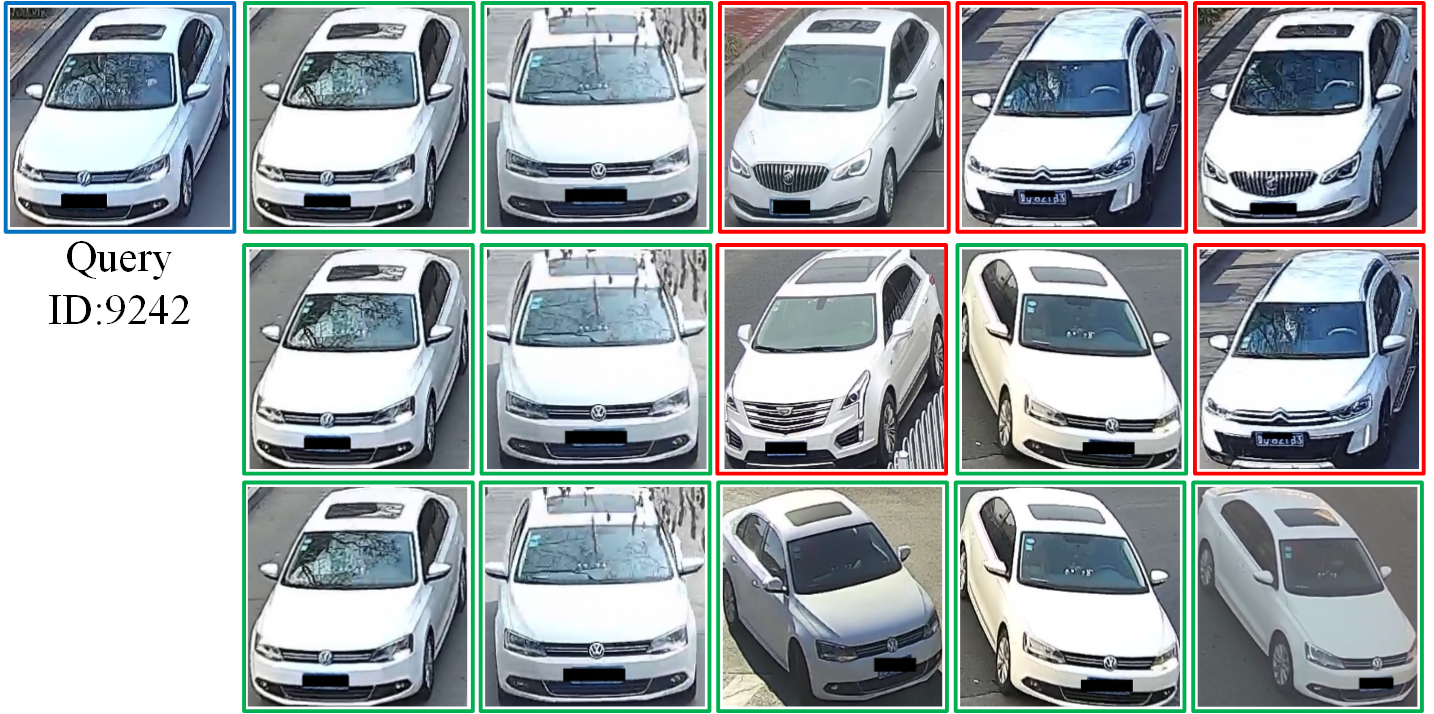}
	%\vspace{-.7cm}
	\caption{Rank-5 visualization examples on VeRi-Wild.
The first, second, and third rows show the top five images returned by Baseline (Global), GiT (No Interactive), and GiT (Interactive), respectively. }
	\label{fig:vis}
%\vspace{-.6cm}
\end{figure}

\begin{figure}[tp]
	\centering
	\includegraphics[width=.8\linewidth]{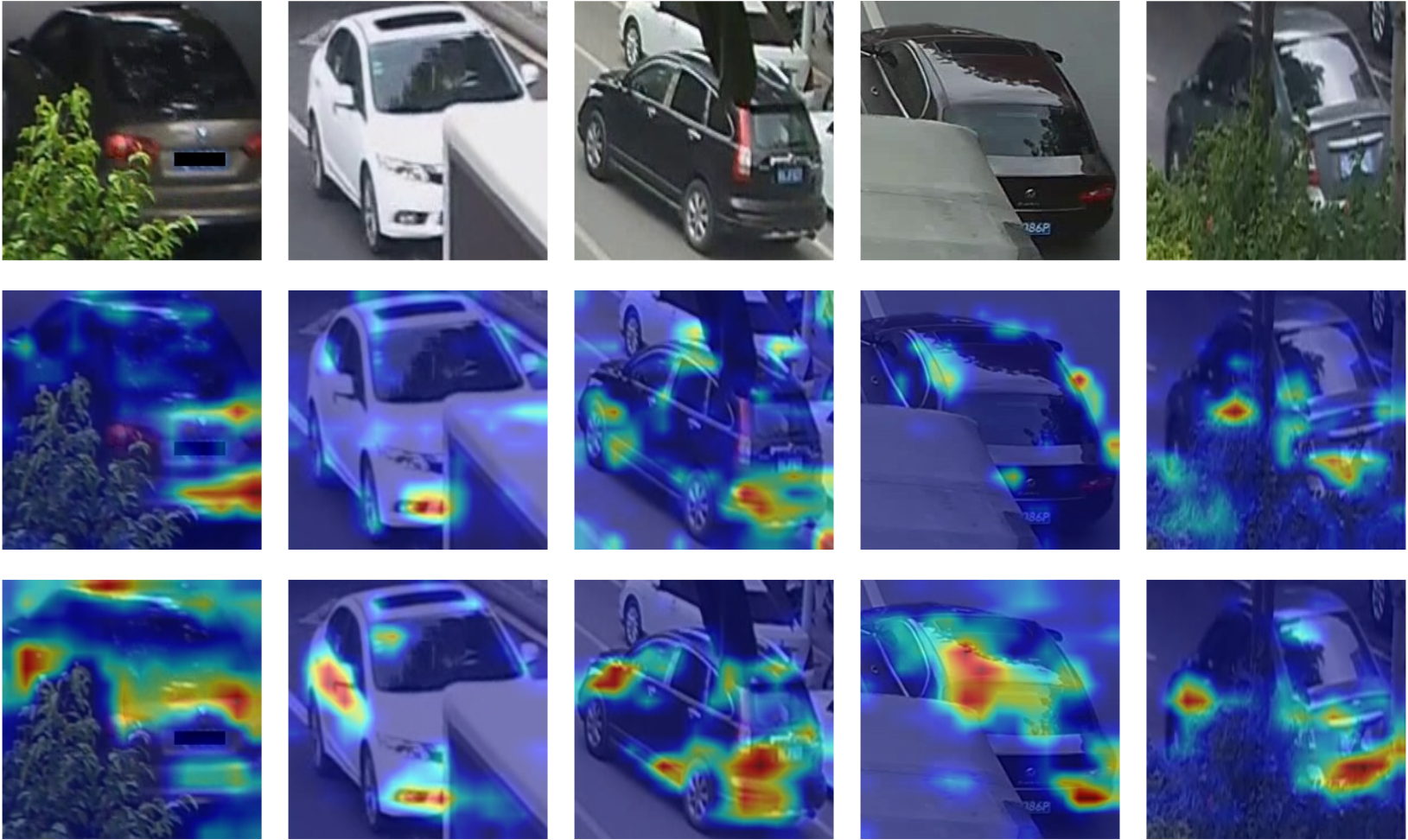}
	%\vspace{-.7cm}
	\caption{The grad class activation maps (Grad-CAM) visualization of attention maps. The first, second, and third rows show the original image, ViT, and GiT, respectively.}
	\label{fig:occlusion}
%\vspace{-.6cm}
\end{figure}

\begin{table*}[tp]
	\renewcommand{\arraystretch}{1.}
	\caption{The performance (\%) comparison among different sampling sizes for feature map in the GiT method.}
	\label{tab:size}
		\vspace{-.1cm}
	\begin{center}
		%\scriptsize
		%\small
		\setlength{\tabcolsep}{4pt}
		\begin{tabular}{lc|cc |cc cc cc |cc cc cc}
			\hline
			%\noalign{\smallskip}
			 \multirow{3}{*}{Names}
			&\multirow{3}{*}{Sampling Sizes}
			& \multicolumn{2}{c|}{\multirow{2}{*}{VeRi776}}
			&\multicolumn{6}{c}{VehicleID}
            &\multicolumn{6}{|c}{VeRi-Wild}\\
			& & &
			& \multicolumn{2}{c}{Test800}
			& \multicolumn{2}{c}{Test1600}
			& \multicolumn{2}{c}{Test2400}
			& \multicolumn{2}{|c}{Test3000}
			& \multicolumn{2}{c}{Test5000}
			& \multicolumn{2}{c}{Test10000}\\
			&
			& R1      &mAP
			& R1      &mAP
			& R1      &mAP
			& R1      &mAP
			& R1      &mAP
			& R1      &mAP
			& R1      &mAP\\
			%\noalign{\smallskip}
			\hline
			GiT1  & (2, 2)$\Rightarrow$(2, 2)$\Rightarrow$(2, 2)
			&96.51 &79.88
			&84.01 &89.25
			&79.44 &85.93
			&76.82 &83.17
			&91.54 &80.79
			&88.80 &74.56
			&84.12 &66.31  \\
	%		PY1 &S5, S4
%			&95.65 &78.91
%			&81.82 &87.93
%			&77.49 &83.96
%			&74.56 &81.12
%			&89.43 &76.52
%			&86.07 &72.42
%			&79.37 &61.25 \\
			GiT2 &(4, 4)$\Rightarrow$(4, 4)$\Rightarrow$(4, 4)
			&96.44 &79.71
			&83.69 &89.13
			&79.30 &85.64
			&76.61 &82.83
			&91.35 &80.42
			&88.67 &74.17
			&83.85 &66.08 \\
			GiT3 & (8, 8)$\Rightarrow$(8, 8)$\Rightarrow$(8, 8)
			&96.33 &79.62
			&83.38 &88.91
			&79.17 &85.35
			&76.36 &82.61
			&91.19 &80.03
			&88.28 &73.64
			&83.19 &65.26 \\
			GiT &(2, 2)$\Rightarrow$(4, 4)$\Rightarrow$(8, 8)
			&\textbf{96.86} &\textbf{80.34}
			&\textbf{84.65} &\textbf{90.12}
			&\textbf{80.52} &\textbf{86.77}
			&\textbf{77.94} &\textbf{84.26}
			&\textbf{92.65} &\textbf{81.76}
			&\textbf{89.92} &\textbf{75.64}
			&\textbf{85.41} &\textbf{67.55} \\
			\hline
		\end{tabular}
	\end{center}
	%\vspace{-.3cm}
\end{table*}

 \subsubsection{\textbf{Role of LCG Module and Transformer Layer Interaction}}
{\color{black}

%We conduct the ablation study of the components in the GiT on the vehicle dataset.
Baseline (Global) denotes the method that uses pure transformer layers without local correlation graph (LCG) modules.
GiT (No Interactive) represents the method applies GiT without the red and blue lines of all GiT blocks in Fig. \ref{fig:framework}.
And similar to Fig. \ref{fig:example} (c), global features and local features are separately supervised with two loss functions.
GiT (Global $\rightarrow$ Local) means the method utilizes GiT without the red lines of all GiT blocks in Fig. \ref{fig:framework}.
GiT (Global $\leftarrow$ Local) denotes the method employs GiT without the blue lines of all GiT blocks in Fig. \ref{fig:framework}.
GiT (Interactive) indicates the full version of GiT, i.e., the red line and blue line are used simultaneously.

\begin{table}[tp]
		\renewcommand{\arraystretch}{1.}
		\caption{Comparison of different backbones on the VeRi776. All methods are implemented on the one same V100 GPU. The FLOPs and AIS respectively denote floating point of operations and average inference speed.}
		\label{tab:flops}
		%\vspace{-.1cm}
		\begin{center}
			%\scriptsize
			\footnotesize
			\setlength{\tabcolsep}{1pt}
			\begin{tabular}{l|c|c|c|c|c}
				\hline
				Methods   & R1  & mAP  & Parameters (M) & FLOPs (G) & {\color{black}AIS (ms/image)}\\ \hline

				ResNet50 \cite{resnet}  & 95.36      & 76.53    & 25.6           & 4.3   &{\color{black}31.47}  \\
				ResNet101 \cite{resnet} & 95.47      & 77.18    & 44.7           & 8.9   &{\color{black}58.26}   \\
				ResNet152  \cite{resnet}& 95.22      & 77.35    & 134.9          & 13.7   &{\color{black}87.61} \\ \hline
				ViT-Tiny \cite{vit} &94.28	&77.84	&29.7	&18.4 & {\color{black}53.38} \\ %\hline
				ViT-Small \cite{vit}& 95.12      & 78.38    & 48.5           & 29.2  &{\color{black}97.24}   \\
				ViT-Base \cite{vit}  & 95.84      & 78.92    & 86.7           & 55.6  &{\color{black}136.17}   \\ \hline
				GiT-Tiny  &96.16	&79.45	&35.1	&18.7  &{\color{black}54.29} \\ %\hline
				GiT-Small  &96.26	&79.72	&53.9	&29.3   &{\color{black}98.38} \\ %\hline
				GiT-Base      & 96.86     & 80.34    & 92.1           & 55.7  &{\color{black}137.63}   \\ \hline
			\end{tabular}
		\end{center}
		%\vspace{-.3cm}
	\end{table}

Firstly, as can be seen in Table \ref{tab:ablation}, the GiT (No Interactive), GiT (Global $\rightarrow$ Local), GiT (Global $\leftarrow$ Local) and GiT have all consistently outperformed Baseline (Global) on two datasets.
This means that we proposed LCG modules that can learn discriminative local features.
The global appearances and local details are complementary for discriminative representations.
Secondly, the result of GiT (Global $\rightarrow$ Local) and GiT (Global $\leftarrow$ Local) are better than GiT (No Interactive) by more than 0.23\% and 0.32\% mAP on the VeRi776 dataset, respectively.
Furthermore, the proposed GiT (Interactive) outperforms both GiT (Global $\rightarrow$ Local) and GiT (Global $\leftarrow$ Local) on two datasets.
For example, compared to the GiT (Global $\rightarrow$ Local) and GiT (Global $\leftarrow$ Local), on VeRi776, the GiT (Interactive) holds a larger 0.52\% and 0.44\% R1 identification rate, respectively.
And on Test2400 of VehicleID, the GiT holds a larger 1.65\% and 1.29 \% mAP, respectively.
{\color{black}
These results showing that either adding transformer’s global features to LCG or adding LCG’s local features to transformer layer is beneficial. Thus, the two-way connection of our GiT fully couples LCG and transformer layer to win the best performance.}
Thirdly, from Fig. \ref{fig:776} and Fig. \ref{fig:2400}, it can be intuitively found that from rank-1 to rank-25, the proposed GiT (Interactive) is better than Baseline (Global) and GiT (No Interactive) on VeRi776 \cite{veri776} and Test2400 of VehicleID \cite{drdl}datasets by a large margin.
These results demonstrate that the interaction of that transformer layer and LCG module can effectively improve the performance of vehicle re-identification.

\subsubsection{\textbf{Visualization}}
In particular, to visualize our proposed GiT (Interactive) method's effectiveness, the rank-5 retrieval results of two example query images from VeRi-Wild are shown in Fig. \ref{fig:vis}.
The first, second, and third rows respectively show the top five images returned by Baseline (Global), GiT (No Interactive), and GiT (Interactive).
Images with blue boxes denote queries.
The images with green boxes and red boxes are correct and incorrect retrieve results, respectively.
It can be seen that Rank1-Rank5 errors of Baseline (Global) and GiT (No Interactive) are often caused by vehicles with high similarity, while the GiT (Interactive) performed well and retrieved the vehicle images with discriminative parts from the query images accurately.
The GiT (No Interactive) method ranks second, while the baseline method has poor results.
This shows that the necessity and effectiveness of global appearances and local information interaction for vehicle re-identification.

{\color{black}
Besides, we visualizing the class activation maps of some examples with occlusion, as shown in Fig. \ref{fig:occlusion}.
From  Fig. \ref{fig:occlusion}, firstly, the grad class activation maps (Grad-CAM) \cite{gradcam} visualized show that both ViT and GiT can focus on non-occluded parts, which illustrates that the visual transformer is a powerful deep learning architecture. Secondly, compared with ViT, one can observe that GiT pays more attention to local differentiated details, e.g., lights and rearview mirrors, which shows that coupling LCG to ViT is beneficial.
}

{\color{black}
\subsubsection{\textbf{Influence of Sampling Strategy in LCG}}

LCG is a key component of GiT method, thus the influence of sampling strategy in LCG is evaluated.  Recalling Table \ref{tab:config}, there are stage1, stage2 and stage3. In Table \ref{tab:size}, the default GiT applies progressive sampling sizes, i.e., LCGs of stage1, stage2 and stage3 use (2, 2), (4, 4) and (8, 8) sampling sizes, respectively. GiT1 method represents the case that always apply the (2, 2) sampling size to LCGs of stage1, stage2 and stage3, i.e., its sampling size does not change among with the depth. Similar to GiT1 method, GiT2 and GiT3 method respectively employ (4, 4) and (8, 8) sampling sizes.

From Table \ref{tab:size}, the GiT method of progressive sampling sizes consistently outperforms GiT1, GiT2 and GiT3 method those of fixed sampling sizes on three different datasets. For example, on the largest Test 10000 of VeRi-Wild , GiT's mAP is 1.24\%, 1.47\% and 2.29\% higher than that of GiT1 method, GiT2 method and GiT3 method, respectively.
On the largest Test2400 of VehicleID, GiT method defeats GiT1 method, GiT2 method and GiT3 method by 1.12\%, 1.33\% and 1.58\% higher R1 identification rates, respectively.
These results imply that the progressive sampling strategy could deal with vehicles scale variations better, thus it is positive strategy for GiT to promote vehicle re-identification performance.

}

\begin{table}[tp]
	\renewcommand{\arraystretch}{1.}
	\caption{The performance (\%) comparison among different scaled GiT on VeRi776. For GiT, ‘H’ means head, ‘D’ is depth, here, GiT-H12-D12 is the default configuration used in this paper.}
	\label{tab:hd}
		%\vspace{-.1cm}
	\begin{center}
		%\scriptsize
		%\small
		\setlength{\tabcolsep}{4pt}

\begin{tabular}{l|c|c|c|c|c}
\hline
Methods           & Heads               & Depths & R1  & mAP  & Parameters (M) \\ \hline
GiT-H3-D12        &   \multirow{3}{*}{3}                 & 12     & 96.16   & 79.45    & 35.1     \\     
GiT-H3-D15        &                     & 15     & 96.22   & 79.61    & 41.9      \\     
GiT-H3-D18        &                     & 18     & 96.47   & 79.89    & 48.6      \\ \hline     
GiT-H6-D12        &   \multirow{3}{*}{6}                    & 12     & 96.26   & 79.72    & 53.9      \\    
GiT-H6-D15        &                     & 15     & 96.51   & 79.94    & 57.3      \\    
GiT-H6-D18        &                     & 18     & 96.73   & 80.11    & 76.8       \\ \hline   
GiT-H12-D12       & \multirow{3}{*}{12}                    & 12     & 96.86   & 80.34    & 92.1        \\   
GiT-H12-D15       &                     & 15     & 97.07   & 80.65    & 113.1        \\  
GiT-H12-D18       &                     & 18     & 97.22   & 81.48    & 134.1        \\ \hline   
\end{tabular}
	\end{center}
	%\vspace{-.3cm}
\end{table}

{\color{black}

\subsubsection{\textbf{Role of Backbones}}
In order to comprehensively the role of backbones, ResNet, ViT, and GiT are  compared at different complexity scales. Following ViT \cite{vit}, {\color{black} we use two critical parameters (i.e., H and D) to control model scales.} H and D denote respectively the number of multi-head self-attention's head and the number of GiT/ViT blocks. In Table \ref{tab:flops}, when the number of depth maintaining 12, the Tiny, Small, and Base respectively denote that those models cost 3, 6, and 12 heads.

From Table \ref{tab:flops}, one can see that ViT and ResNet hold comparable R1 and mAP.
{\color{black}For example, the ResNet-101's R1 and ViT-Small's R1 are 95.47\% and 95.12\% performance on VeRi776, respectively.}These comparisons clearly show that GiT’s improvement is not caused by using better ViT backbones.

\subsubsection{\textbf{Complexity Analysis}}
We compare different CNN and Transformer models, as shown in Table \ref{tab:flops}. The FLOPs denotes floating point of operations. Both GiT and ViT are inferior to that of ResNet, as shown in Table \ref{tab:flops}. But, even coupling graph neural networks, both FLOPs and parameter count of GiT are comparable to these of ViT. Especially, GiT-Tiny defeats ViT-Base in terms of complexity and accuracy. {\color{black}
Besides, for an intuitive analysis, the average inference speed (AIS) is applied, which is the average time for feature extraction of each image. Similar to the FLOPs comparison, in terms of AIS, both GiT and ViT are inferior to that of ResNet, as shown in Table \ref{tab:flops}. However, GiT still are comparable to ViT, even GiT extra introduces graph neural networks (GNN).} Therefore, our GiT is an efficient transformer, although its efficiency is weaker than ResNet.

\subsubsection{\textbf{Sensitivity Analysis}}
For Table \ref{tab:hd}, given the GiT-H3-D12 model as an example, it uses 3-head self-attention and 12 GiT blocks. From Table \ref{tab:hd}, one can see that either an increase in H or D increases the number of GiT’s parameters and improves performance. For example, when H=12, D increases from 12 to 18, parameters only raise from 92.1M to 134.1M, an increase of 45\%. But when D=12, H increases from 3 to 12, parameters significantly expand from 35.1M to 92.1M, an increase of 162.3\%.}}

{\color {black}
	\begin{table}[tp]
		\renewcommand{\arraystretch}{1.0}
		\caption{The Performance (\%) Comparison with state-of-the-art methods on Market-1501 and MSMT17.
		}\label{tab:person}
		%\vspace{-.3cm}
		\begin{center}
			%\footnotesize
			%\setlength{\tabcolsep}{2pt}
			\begin{tabular}{l|cc|cc}
				\hline
				\multirow{2}{*}{Methods} & \multicolumn{2}{c|}{Market-1501 \cite{market}} & \multicolumn{2}{c}{MSMT17 \cite{msmt17}} \\ \cline{2-5}
				& \multicolumn{1}{c}{R1}        & mAP       & \multicolumn{1}{c}{R1}      & mAP    \\ \hline
				PCB \cite{pcb}            & \multicolumn{1}{c}{93.8}      & 81.6      & \multicolumn{1}{c}{68.2}    & 40.4   \\
				OSNet \cite{osnet}           & \multicolumn{1}{c}{94.8}      & 84.9      & \multicolumn{1}{c}{78.7}    & 52.9   \\
				DG-Net \cite{dgnet}          & \multicolumn{1}{c}{94.8}      & 86.0      & \multicolumn{1}{c}{77.2}    & 52.3   \\
				MGN \cite{mgn}             & \multicolumn{1}{c}{95.7}      & 86.9      & \multicolumn{1}{c}{76.9}    & 52.1   \\
				RGA-SC \cite{rga-sc}          & \multicolumn{1}{c}{96.1}      & 88.4      & \multicolumn{1}{c}{80.3}    & 57.5   \\
				SAN \cite{san}             & \multicolumn{1}{c}{96.1}      & 88.0      & \multicolumn{1}{c}{79.2}    & 55.7   \\
				SCSN \cite{scsn}            & \multicolumn{1}{c}{95.7}      & 88.5      & \multicolumn{1}{c}{83.8}    & 58.5   \\
				ABDNet \cite{abdnet}          & \multicolumn{1}{c}{95.6}      & 88.3      & \multicolumn{1}{c}{82.3}    & 60.8   \\ \hline
				ViT \cite{vit}                 & \multicolumn{1}{c}{94.9}      & 87.0      & \multicolumn{1}{c}{82.1}    & 61.5   \\
				GiT                      & \multicolumn{1}{c}{95.7}      & 88.9      & \multicolumn{1}{c}{85.6}    & 64.8   \\ \hline
			\end{tabular}
		\end{center}
		%\vspace{-.3cm}
	\end{table}
	\subsection{Generalization to Person Re-identification}
	As shown in Table \ref{tab:person}, our proposed GiT is compared with state-of-the-art methods on person re-identification datasets, i.e., Market-1501 \cite{market} and MSMT17 \cite{msmt17} datasets.
{\color{black} On Market-1501,} our GiT obtains the highest mAP (i.e., 88.9\%), although the R1 is a bit lower than RGA-SC \cite{rga-sc} and SAN [23] \cite{san}. On the larger MSMT17 dataset, both RGA-SC \cite{rga-sc} and SAN [23] \cite{san} loss dominance, but or GiT still is strong to win best R1 and mAP. These results show that our GiT is general to person re-identification.

}

\section{Conclusion}\label{sec:con}
{\color{black}
In this paper, we propose a graph interactive transformer (GiT) method for vehicle re-identification. The GiT method couples graphs and transformers to explore the interaction between local features and global features, resulting in effective cooperation between local and global features. We also design a new local correlation graph (LCG) module for learning local features within patches.
We construct extensive experiments to analyze our GiT method, including: (1) Role of LCG module and transformer layer interaction.
(2)  Visualization;
(3) Influence of sampling strategy in LCG module;
(4)  Role of backbones;
(5)   Complexity analysis;
 (6)  Sensitivity analysis.

}
\ifCLASSOPTIONcaptionsoff
  \newpage
\fi

\bibliographystyle{IEEEtran}
\small{
\bibliography{reffullv2}

}

\end{document}